# An Adam-adjusting-antennae BAS Algorithm for Refining Latent Factors

Yuanyi Liu
School of Cyber Science and Technology
Beihang University
Beijing, China
YuanyiLiu@buaa.edu.cn

Jia Chen
School of Cyber Science and Technology
Beihang University
Beijing, China
chenjia@buaa.edu.cn

Di Wu*
College of Computer and Information Science,
Southwest University,
Chongqing 400715, China
wudi.cigit@gmail.com

*Abstract*—Extracting the latent information in high-dimensional and incomplete matrices is an important and challenging issue. The Latent Factor Analysis (LFA) model can well handle the high-dimensional matrices analysis. Recently, Particle Swarm Optimization (PSO)-incorporated LFA models have been proposed to tune the hyper-parameters adaptively with high efficiency. However, the incorporation of PSO causes the premature problem. To address this issue, we propose a sequential Adam-adjusting-antennae BAS ($A^2$BAS) optimization algorithm, which refines the latent factors obtained by the PSO-incorporated LFA model. The $A^2$BAS algorithm consists of two sub-algorithms. First, we design an improved BAS algorithm which adjusts beetles' antennae and step-size with Adam; Second, we implement the improved BAS algorithm to optimize all the row and column latent factors sequentially. With experimental results on two real high-dimensional matrices, we demonstrate that our algorithm can effectively solve the premature convergence issue.

*Keywords*—High-dimensional and Incomplete (HDI) Matrix, Latent Factor Analysis (LFA), BAS algorithm, ADAM algorithm, Industrial Application Big Data.

## I. INTRODUCTION

High-Dimensional and Incomplete (HDI) matrices, which contain large amounts of extremely incomplete data, are adopted widely in social network services[1-5], recommender systems[6-9], and big-data related applications[10-14]. With the rapid expansion of these services, the HDI matrices have been accumulated rapidly, containing more and more useful information. Extracting the useful latent factors from HDI matrices becomes a challenging research issue.

Among the state-of-the-art approaches for analyzing the HDI matrices, the Latent Factor Analysis (LFA) model has outstanding performance because of its high scalability and efficiency[15-23]. The LFA model maps the row and column entities of a target HDI matrix into a low-dimensional latent factor space to build its low rank approximation based on its known data. It takes four steps to build the model: (1) latent factor mapping, (2) low-rank approximation, (3) learning objective construction, and (4) iterative optimization. Among these four steps, the optimization process influences both the analytical accuracy and efficiency of HDI matrices.

The popular optimization algorithms include Stochastic Gradient Descent (SGD) algorithm[24], AdaGrad[25], AdaDelta[26] and Adam[27]. The SGD algorithm is implemented widely in LFA model due to its superiority in convergence speed. In order to solve the hyper-parameters manually setting problem, AdaGrad, AdaDelta and Adam, have been proposed. AdaGrad adjusts its learning rate via calculating the sum squares of gradients in former iterations. AdaDelta adjusts the learning rate based on the decaying average of all the past squared gradients. Adam incorporates the first-order and second-order gradients together to optimize the objective function. All these learning rate self-adapting algorithms obtain higher accuracy with additional time cost.

In addition to improving the efficiency of optimization algorithms above, Luo et al.[20] incorporate the Particle Swarm Optimization (PSO) algorithm, which is a kind of swarm intelligence algorithm, into LFA model to improve its optimization process with considerable time cost. The representative one is the Position-transitional Particle Swarm Optimization-based LFA (PLFA) model, which adopts the dynamical PSO algorithm to adjust the LFA's learning rate adaptively [18]. These PSO-incorporated LFA models prove that swarm intelligence algorithms can boost the LFA model's optimization efficiently. However, although these PSO-related LFA models can accelerate convergency while adjusting the learning rate adaptively, they bring the premature convergence problem. We adopt the Beetle Antennae Search (BAS) algorithm to address this issue.

Beetle Antennae Search (BAS) algorithm is an innovative swarm intelligence algorithm, which simulates the beetles' step-by-step foraging process to find out the global optimum. The beetles jump out the local optimum quite easily. Thereby the convergence speed is fast, which promotes the wide application of BAS in large-scale data processing [28]. Recently, many researchers have improved the BAS model and adopt it into many application fields. Beetle Swarm Optimization (BSO) algorithm incorporates PSO into BAS algorithm and constructs the cluster to avoid falling into local optimum [29]. Beetle Colony Optimization (BCO) algorithm adopts a feedback-step update strategy to adjust the beetle's searching step size [30]. BAS-ADAM algorithm adopts Adam to adjust the beetle's searching step-size adaptively [31]. We improve the BAS-Adam algorithm and adopt it to refine the latent factors for an LFA model.

This work is supported in part by the National Natural Science Foundation of China under grant 62176070. (Corresponding author: D. Wu., Y. Liu and J. Chen are co-first authors of this paper.)

In this paper, we propose an Adam-adjusting-antennae BAS (A²BAS) optimization algorithm to refine the latent factor calculated by a PLFA model, our method can resolve the premature problem in the PSO-based LFA model. Our main contributions are listed as follows:

a) An A²BAS algorithm. The A²BAS algorithm adjusts beetles' antennae with Adam, and then update the beetle's position with a mapping function of the antennae. The proposed algorithm optimizes a single latent factor vector in a target HDI matrix.

b) A sequential A²BAS-based optimization algorithm. we apply a sequential A²BAS-based algorithm to refine the latent factor achieved by a PLFA model.

Section II describes some important concepts and algorithms. Section III proposes our innovative methods. Section IV provides experimental data and analyzes the empirical results. Finally, Section V concludes this paper and plans for future work.

## II. PRELIMINARIES

In this section, we give clear definitions of a LFA model, a PLFA model and a BAS-ADAM model.

### A. Problem Statement

Table I lists the symbols appear in our paper. We introduce the definition of an LFA model to analyze HDI matrices.

**Definition 1. An LFA model.** An LFA model adopts two low-dimensional latent factor matrices, i.e., $P$ and $Q$, to construct a latent factor estimation $\hat{R}=PQ^T$ for $R$. $P^{|E|\times f}$ and $Q^{|U|\times f}$ denote the latent feature matrices of two large entities $E$ and $U$, respectively. And $\Lambda$ denotes the known dataset of $R$ to construct the objective function.

To measure the differences between $\hat{R}$ and $R$, we adopt Euclidean distance as the objective function, and add $L2$-norm-based regularization to avoid overfitting. The objective function is defined as:

$$\varepsilon(P,Q,\mathbf{c},\mathbf{d}) = \frac{1}{2}\sum_{r_{e,u}\in\Lambda}\left(r_{e,u}-\sum_{k=1}^{f}p_{e,k}q_{k,u}-c_e-d_u\right)^2 + \frac{\lambda}{2}\sum_{r_{e,u}\in\Lambda}\left(\sum_{k=1}^{f}p_{e,k}^2+\sum_{k=1}^{f}q_{k,u}^2+c_e^2+d_u^2\right), \quad (1)$$

where $\lambda$ denotes the regularization coefficient, $c_e$ and $d_u$ indicate linear biases for $e$ and $u$. $p_{e,k}$ and $q_{k,u}$ denote the single element in $P$ and $Q$, respectively.

TABLE I. ADOPTED SYMBOLS WITH THEIR DESCRIPTION

| Symbol | Description |
|---|---|
| $E, U$ | Two large entity sets with some relationships. |
| $e, u$ | An entity instance in $E$ and $U$, respectively. |
| $R, \hat{R}$ | An HDI matrix containing the relationships between $E$ and $U$, and its approximate matrix. |
| $r_{e,u}, \hat{r}_{e,u}$ | The relationship strength between $e$ and $u$. |
| $\Lambda, \Gamma$ | Known and unknown data sets of $R$. |
| $P, Q$ | Latent factor matrices for $E$ and $U$, respectively. |
| $\mathbf{p}_e, \mathbf{q}_u$ | The $e$-th row vector of $P$, and the $u$-th row vector of $Q$. |
| $F(\cdot)$ | Fitness function with 2-norm. |
| $p_{e,k}, q_{k,u}$ | A latent factor element of $\mathbf{p}_e$ and $\mathbf{q}_u$, respectively. |
| $\mathbf{c}, \mathbf{d}$ | Linear biases for $E$ and $U$, respectively. |
| $c_e, d_u$ | A linear bias for $e$ and $u$, respectively. |
| $f$ | Dimension of the latent factor space. |
| $\varepsilon$ | Loss function measuring the difference between $\hat{R}$ and $\Lambda$. |
| $\lambda$ | Regulation parameter for an LFA model. |
| $\eta$ | Learning rate for an SGD-based LFA model. |
| $\omega$ | A balancing constant in PSO's velocity formula. |
| $\gamma_1, \gamma_2$ | Two cognitive and social coefficients. |
| $r_1, r_2$ | Two uniform random variables between 0 and 1. |
| $\rho$ | A scaling factor of infected position component. |
| $\mathbf{vp}_k(t), \mathbf{xp}_k(t)$ | The $k$-th particle's velocity and position at the $t$-th iteration. |
| $X, \mathbf{x}_i$ | A set of antennae locations, and a single element in it. |
| $X_l, X_h$ | Two subsets of $X$, containing $N$ lowest and highest antennae. |
| $\mathbf{xc}_{tl}, \mathbf{xc}_{th}$ | The centroids of sets $X_l$ and $X_h$. |
| $\nabla_t$ | Gradient estimation of current position. |
| $\mathbf{m}_t, \mathbf{v}_t$ | First and second moment of the estimated gradient. |
| $\mathbf{x}_{new}$ | A beetle's updated position. |
| $\delta_t, \delta_0$ | Beetle's step size at $t$ iteration, and the initial step size. |
| $\beta_1, \beta_2, \alpha$ | Parameters used in Adam algorithm. |
| $\mathbf{m}_r(t), \mathbf{m}_l(t)$ | First moment for the right and left antennae. |
| $\hat{\mathbf{m}}_r(t), \hat{\mathbf{m}}_l(t)$ | Unbiased first moment for the right and left antennae. |
| $\mathbf{v}_r(t), \mathbf{v}_l(t)$ | Second moments for the right and left antennae |
| $\hat{\mathbf{v}}_r(t), \hat{\mathbf{v}}_l(t)$ | Unbiased second moments for the right and left antennae. |
| $\mathbf{x}_r(t), \mathbf{x}_l(t)$ | The right and left antennae positions. |
| $\mathbf{x}_{temp}(t)$ | A beetle's optimum position at the $t$-th iteration. |
| $al$ | A beetle's antennae length. |
| $\mathbf{x}_{best}, F_{best}$ | Global optimum position and fitness function value. |

SGD is a popular algorithm to optimize an LFA model. During its step-by-step optimization process, the latent factors in an iteration can be updated as:

$$\forall r_{e,u} \in \Lambda, k \in [1, f]: \begin{cases} p_{e,k}^m \leftarrow p_{e,k}^{m-1} + \eta \cdot (\Delta_{e,u}^{m-1} \cdot q_{k,u}^{m-1} - \lambda \cdot p_{e,k}^{m-1}), \ q_{k,u}^m \leftarrow q_{k,u}^{m-1} + \eta \cdot (\Delta_{e,u}^{m-1} \cdot p_{e,k}^{m-1} - \lambda \cdot q_{k,u}^{m-1}), \\ c_e^m \leftarrow c_e^{m-1} \cdot (1 + \eta \cdot \Delta_{e,u}^{m-1} - \lambda \cdot \eta), \ d_u^m \leftarrow d_u^{m-1} \cdot (1 + \eta \cdot \Delta_{e,u}^{m-1} - \lambda \cdot \eta), \end{cases} \quad (2)$$

where $\Delta_{e,u} = r_{e,u} - \sum_{k=1}^{f} p_{e,k} q_{k,u} - c_e - d_u$, $\eta$ denotes the constant learning rate, $m$ denotes the iteration number.

*B. A PLFA Model*

Through adopting a P²SO algorithm to adjust an SGD-based LFA model's learning rate, a PLFA model achieves high latent factor analysis accuracy. P²SO algorithm generates a particle swarm, in which each particle represents a learning rate and more position information is injected into each particle's evolution. the $k$-th particle's velocity and position update formulas at the $t$-th iteration are represented as:

$$\begin{cases} \mathbf{vp}_k(t) = \omega \mathbf{vp}_k(t-1) + \gamma_1 r_1 (\hat{\mathbf{p}}_k(t-1) - \mathbf{xp}_k(t-1)) + \gamma_2 r_2 (\hat{\mathbf{g}}(t-1) - \mathbf{xp}_k(t-1)) + \rho(\gamma_1 r_1 + \gamma_2 r_2)(\mathbf{xp}_k(t-1) - \mathbf{vp}_k(t-1)), \\ \mathbf{xp}_k(t) = \mathbf{xp}_k(t-1) + \mathbf{vp}_k(t), \end{cases} \quad (3)$$

where $\omega$ is the non-negative constant, $\gamma_1$ and $\gamma_2$ are cognitive and social coefficients, $r_1$ and $r_2$ are two uniform random variables in the range of [0, 1], $\rho$ is a scaling factor of injected position component.

The P²SO algorithm updates all the particles until the swarm convergences. Then PLFA adopts the P²SO's global best position as a new learning rate.

*C. A BAS-ADAM Algorithm*

The BAS-ADAM algorithm incorporates Adam into BAS algorithm to adjust the beetle's searching-step size. First, BAS-ADAM constructs two subsets $X_l$ and $X_h$ of all the randomly generated antennae adjacent to a beetle $\mathbf{x}_t$. The $N$ lowest and $N$ highest antennae are sorted with fitness function and stored into $X_l$ and $X_h$, respectively. The centroids $\mathbf{xc}_{tl}$ and $\mathbf{xc}_{th}$ of subsets $X_l$ and $X_h$ are calculated as:

$$\forall \mathbf{x}_t \in X: \mathbf{xc}_{tl} = \sum_{\mathbf{x}_{tl} \in X_l} \frac{\mathbf{x}_{tl}}{N}, \mathbf{xc}_{th} = \sum_{\mathbf{x}_{th} \in X_h} \frac{\mathbf{x}_{th}}{N} \quad (4)$$

Thereby, the gradient $\nabla_t$ is calculated as difference between $\mathbf{xc}_{th}$ and $\mathbf{xc}_{tl}$. And the first moment $\mathbf{m}_t$ and second moment $\mathbf{v}_t$ of $\nabla_t$ can be calculated as:

$$\begin{cases} \mathbf{m}_t = \beta_1 \mathbf{m}_{t-1} + (1 - \beta_1) sign(F(\mathbf{xc}_{th}) - F(\mathbf{xc}_{tl})) \nabla_t, \\ \mathbf{v}_t = \beta_2 \mathbf{v}_{t-1} + (1 - \beta_2) \nabla_t^2, \end{cases} \quad (5)$$

where $\beta_1$ and $\beta_2$ are the hyper-parameters in Adam, $F(.)$ is the fitness function, $sign(.)$ is the direction of the increasing fitness function. Because $\mathbf{m}_t$ and $\mathbf{v}_t$ are biased toward 0, the bias-corrected $\hat{\mathbf{m}}_t$ and $\hat{\mathbf{v}}_t$ are calculated and adopted to update $\mathbf{x}_t$, the new position $\mathbf{x}_{new}$ can be defined as:

$$\mathbf{x}_{new} = \mathbf{x}_t + \delta_t \frac{\hat{\mathbf{m}}_t}{\sqrt{\hat{\mathbf{v}}_t} + \varepsilon}, \quad (6)$$

where $\delta_t$ indicates the step length at $t$-th iteration, whose value decreases linearly with the iteration rounds increases, $\varepsilon$ is the parameter set to avoid denominator becoming zero. Then, the fitness values of $\mathbf{x}_{new}$ and $\mathbf{x}_t$ are compared and the smaller one is chosen as beetle's new position.

BAS-ADAM algorithm achieves higher performance with less iterations compared with original BAS and PSO. Thereby, we improve it for the HDI matrices processing.

### III. METHODOLOGY

*A. An A²BAS Algorithm for the Single Latent Factor*

The A²BAS algorithm constructs a beetle to optimize a specific latent factor, then adjusts the beetle's antennae through Adam algorithm and updates its position with the antennae iteratively.

**Initialization of a beetle.** Each beetle is initialized as a single latent factor component for entity one or entity two. More specifically, for a beetle representing a single latent factor of entity one, its initial position $\mathbf{x}_e(0)$ is given as:

$$\mathbf{x}_e(0) = [\mathbf{p}_e, c_e], \forall e \in E, \quad (7)$$

where $[\mathbf{p}_e, c_e]$ denotes the $(f+1)$-dimensional latent factor for the specific entity $e$, and for a beetle representing a single latent factor in entity two, its initial position $\mathbf{x}_u(0)$ can be formulated as:

$$\mathbf{x}_u(0) = [\mathbf{q}_u, d_u], \forall u \in U, \quad (8)$$

The best location $\mathbf{x}_{best}$ of a beetle in entity one is initialized as $\mathbf{x}_e(0)$, and $\mathbf{x}_{best}$ is initialized as $\mathbf{x}_u(0)$ for a beetle in entity two. Then, we randomly generate a beetle's $f+1$ dimensional initial antennae direction $\mathbf{dr}(0)$, and initialize left antennae and right

antennae for a beetle as:

$$\begin{cases} \mathbf{x}_r(0) = x(0) + \mathbf{dr}(0), \\ \mathbf{x}_l(0) = x(0) - \mathbf{dr}(0). \end{cases} \quad (9)$$

**Construct the fitness function.** To find out the best position for optimizing the latent factor in entity one, we construct the fitness function with the objective function in (1). Fitness functions are formulated as:

$$\begin{aligned} F_1(x^e(t)) &= \frac{1}{2} \sum_{r_{e,u} \in \Lambda(e)} \left(r_{e,u} - \mathbf{p}_e q_u - c_e - d_u\right)^2 + \frac{\lambda}{2}\left(\|\mathbf{p}_e\|^2 + \|c_e\|^2\right), \\ F_2(x^e(t)) &= \sum_{r_{e,u} \in \Lambda(e)} \left|r_{e,u} - \mathbf{p}_e q_u - c_e - d_u\right| + \left(|\mathbf{p}_e| + |c_e|\right), \end{aligned} \quad (10)$$

where $F_1(\cdot)$ is the fitness function for RMSE, and $F_2(\cdot)$ is the fitness function for MAE. $|\cdot|$ indicates the absolute value. Similar to (10), we also construct the fitness function for optimizing a beetle which represents a latent factor in entity two. The formular can be given as:

$$\begin{aligned} F_1(x^u(t)) &= \frac{1}{2} \sum_{r_{e,u} \in \Lambda(u)} \left(r_{e,u} - p_e \mathbf{q}_u - c_e - d_u\right)^2 + \frac{\lambda}{2}\left(\|\mathbf{q}_u\|^2 + \|d_u\|^2\right), \\ F_2(x^u(t)) &= \sum_{r_{e,u} \in \Lambda(u)} \left|r_{e,u} - p_e \mathbf{q}_u - c_e - d_u\right| + \left(|\mathbf{q}_u| + |d_u|\right). \end{aligned} \quad (11)$$

**Update beetle's antennae with Adam.** First, for the $t$-th iteration, we randomly generate a $f+1$ dimensional antennae direction vector $\mathbf{dr}(t)$,

$$\mathbf{dr}(t) = \frac{rand(f+1)}{\|rand(f+1)\|}, \quad (12)$$

where $rand(\cdot)$ indicates a random function, which returns a $f+1$ dimensional vector between 0 and 1. Then, we apply Adam algorithm to update beetle's antennae. The first moment of the right and left antennae are formulated as:

$$\begin{cases} \mathbf{m}_r(t) = \beta_1 \mathbf{m}_r(t-1) + (1-\beta_1)\mathbf{dr}(t), \\ \mathbf{m}_l(t) = \beta_1 \mathbf{m}_l(t-1) - (1-\beta_1)\mathbf{dr}(t), \end{cases} \quad (13)$$

where $\beta_1$ is a hyper-parameter balancing the historical first moment vector and the instant antennae direction. Their unbiased correction terms can be represented as:

$$\begin{cases} \hat{\mathbf{m}}_r(t) = \mathbf{m}_r(t)/(1-\beta_1^t), \\ \hat{\mathbf{m}}_l(t) = \mathbf{m}_l(t)/(1-\beta_1^t), \end{cases} \quad (14)$$

where $\beta_1^t$ denotes $t$-th exponent of $\beta_1$. The second moment of right and left antennae are given as:

$$\begin{cases} \mathbf{v}_r(t) = \beta_2 \mathbf{v}_r(t-1) + (1-\beta_2)\mathbf{dr}(t) \cdot \mathbf{dr}(t), \\ \mathbf{v}_l(t) = (1-\beta_2)\mathbf{dr}(t) \cdot \mathbf{dr}(t) - \beta_2 \mathbf{v}_l(t-1), \end{cases} \quad (15)$$

where $\beta_2$ is a hyper-parameter balancing the historical second moment vector and the instant antennae direction. And their unbiased second moment terms are given as:

$$\begin{cases} \hat{\mathbf{v}}_r(t) = \mathbf{v}_r(t)/(1-\beta_2^t), \\ \hat{\mathbf{v}}_l(t) = \mathbf{v}_l(t)/(1-\beta_2^t). \end{cases} \quad (16)$$

Based on (12)-(16), we calculate the right antenna $\mathbf{x}_r(t)$ and left antenna $\mathbf{x}_l(t)$ as follows:

$$\begin{cases} \mathbf{x}_r(t) = \mathbf{x}_{best}(t-1) + \alpha \cdot al(t-1) \cdot \dfrac{\hat{\mathbf{m}}_r(t)}{\sqrt{\hat{\mathbf{v}}_r(t)} + \varepsilon}, \\ \mathbf{x}_l(t) = \mathbf{x}_{best}(t-1) - \alpha \cdot al(t-1) \cdot \dfrac{\hat{\mathbf{m}}_l(t)}{\sqrt{\hat{\mathbf{v}}_l(t)} + \varepsilon}, \end{cases} \quad (17)$$

where $\alpha$ is a hyper-parameter balancing the historical beetle's position and the position alteration. $\varepsilon$ is a tiny constant to avoid denominator equaling 0. $al$ indicates antennae length, which effect searching field diameter.

**Update beetle's position.** In order to guarantee fitness value decreases after iteration, fitness value of $\mathbf{x}_r(t)$ and $\mathbf{x}_l(t)$ are calculated and compared. Antennae with smaller fitness value will be chosen as beetle's moving direction. Beetle's temporarily optimum position $\mathbf{x}_{temp}(t)$ is updated as:

$$\mathbf{x}_{temp}(t) = \begin{cases} \mathbf{x}_r(t), & \text{if } F(\mathbf{x}_r(t)) < F(\mathbf{x}_l(t)), \\ \mathbf{x}_l(t), & \text{if } F(\mathbf{x}_r(t)) \geq F(\mathbf{x}_l(t)), \end{cases} \quad (18)$$

Then, we compare value between $F(\mathbf{x}_{temp}(t))$ and global best value $F_{best}$. If temporary optimum value is smaller than $F_{best}$, we replace $F_{best}$. Besides, we update antennae length $al$ to shrink searching domain in next generation.

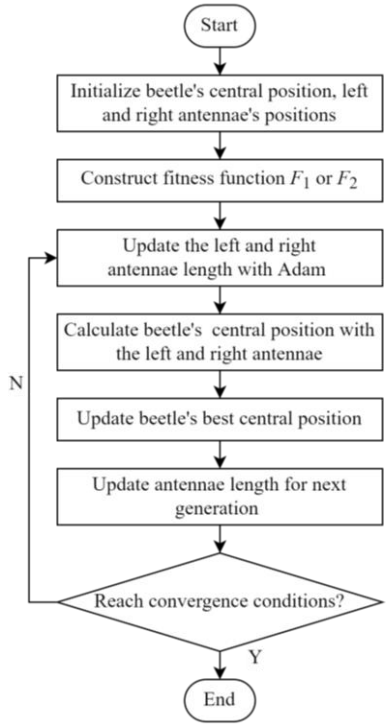

Fig.1 Flow Chart of the A²BAS Algorithm

The whole optimization process continues until the $F_{best}$ convergences. Figure.1 shows the procedure of a single LF- based A²BAS algorithm.

*B. A Sequential A²BAS Optimization Algorithm*

Based on the A²BAS algorithm for a single latent factor in previous subsection, we design a sequential A²BAS algorithm for all the latent factors in the target HDI matrix.

First, the latent factors obtained by a PLFA model are divided into $|E|$ row groups and $|U|$ column groups. Each row group contains a latent factor pair [$\mathbf{p}_e$, $c_e$], and each column group contains a latent factor pair [$\mathbf{q}_u$, $d_u$]. Second, for each [$\mathbf{p}_e$, $c_e$], $\forall e \leq |E|$, we implement A²BAS algorithm while keeping other parameters fixed. Each updated latent factor pair are transferred back to $P$, $C$. Third, for each [$\mathbf{q}_u$, $d_u$], $\forall u \leq |U|$, we implement A²BAS algorithm while keeping other parameters fixed. Each updated latent factor pair are transferred back to $Q$, $D$. Finally, calculate RMSE or MAE value on validation sets, and compare value between two adjacent optimization rounds. If new RMSE/MAE value is lower than the former round, update the model.

IV. EXPERIMENTAL RESULTS AND ANALYSIS

*A. General Settings*

**Evaluation Protocol.** In order to compare the prediction accuracy of our proposed model with others, we choose RMSE and MAE to measure the models' accuracy. To compare the models' efficiency, we choose iteration counts and time cost as criterion. All experiments are performed on a workstation with a 3.20 GHz AMD Ryze 7 CPU and 16 GB RAM. The formulas of RMSE and MAE are given as:

$$RMSE = \sqrt{\frac{1}{|\psi|}\left(\sum_{r_{u,i} \in \Lambda}(r_{u,i} - \hat{r}_{u,i})^2\right)}, MAE = \frac{1}{|\psi|}\left(\sum_{r_{u,i} \in \Lambda}|r_{u,i} - \hat{r}_{u,i}|\right).$$

**Dataset.** We adopt two datasets ML10M [32] and Flixster [33] as D1 and D2. We divide each dataset into three parts, which holds 70%-10%-20% train-validation-test settings.

**Model Settings.** We adopt Adam model, PLFA model, and HPL model as benchmark. The dimension of LF space $f$ is set as 20. The regularization coefficient $\lambda$ and learning rate $\eta$ are tuned in advanced for each dataset, which are set as [0.03, 0.03] and [0.015, 0.02] for D1 and D2.

In order to ensure the consistency of experimental results, we set the hyper-parameters in PLFA model as previous settings in [19] and [34]. Besides, we set the Adam-related hyper-parameters same as in [27].

**Experiment Design.** We design two experiments to investigate the performance of A²BAS algorithm. First, we investigate how the various initial antennae length influences the accuracy and efficiency of A²BAS algorithm. Second, we compare the A²BAS algorithm with other benchmark models.

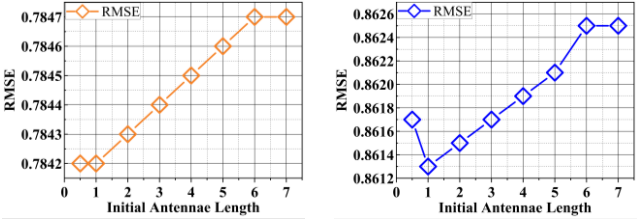

(a) RMSE on D1  (b) RMSE on D2
Fig. 2. RMSE of A²BAS as length increases.

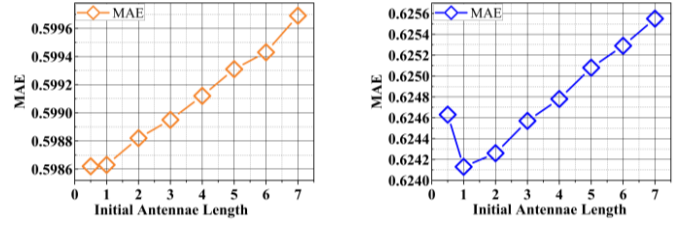

(a) MAE on D1  (b) MAE on D2
Fig. 3. MAE of A²BAS as antennae length increases.

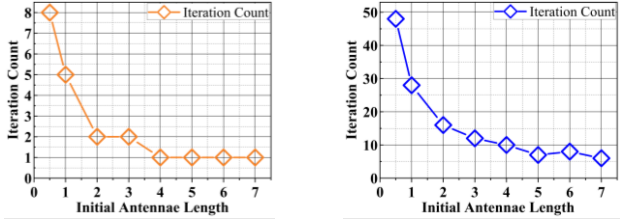

(a) Time Cost on D1  (b) Time Cost on D2
Fig. 4. Converging iterations of A²BAS as antennae length increases.

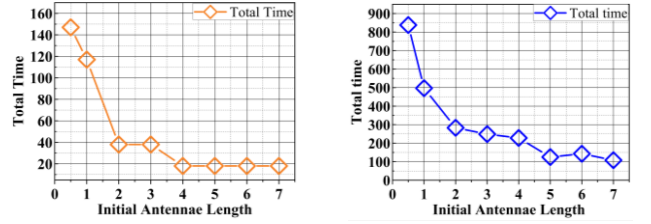

(a) Iterations to Converge on D1  (b) Iterations to Converge on D2
Fig. 5. Time cost of A²BAS as antennae length increases.

*B. Performance comparison*

*1) Comparison of antennae length*

For both D1 and D2, we conduct experiments with the antennae length changing from 0.1 to 13. Observing the results, we set the various initial antennae lengths for further experiments as al_set = [0.5, 1, 2, 3, 4, 5, 6, 7]. Thereby, we depict Fig.2 to Fig.5 to demonstrate the performance changes on D1 and D2 with various initial antennae lengths. We conclude the following findings:

a) The prediction accuracy fluctuates quite little with various antennae lengths, and it becomes larger slowly when antennae length increases. According to Fig.2, the standard deviations of A²BAS's RMSE on D1 and D2 are 2.07E-4 and 4.41E-4, respectively. And as shown in Fig.3, the standard deviations of A²BAS's MAE on D1 and D2 are 3.87E-4 and 4.93E-4. The differences in RMSE and MAE on D1 and D2 are rather small as initial antennae length varies.

b) The convergence process of A²BAS is fast. It can be seen from Fig.4 (a) that, A²BAS optimization process takes 2 iterations when initial antennae length=2.0, and only takes 1 iteration when initial antennae length⩾4.0. It should be emphasized that when initial antennae length is sufficiently large, iteration round will keep stable at a low scale.

c) The total time consumption of A²BAS decreases with initial antennae length increases. For example, in Fig.5 (a), as initial antennae length increases from 0.5 to 5.0, total time cost drops monotonically from 147 to 18. It is worth noting that, total time consumption decreases almost linearly as iteration round decreases, and when iteration round remains basically the same, total time cost also keeps stable.

Based on the above analysis, we conclude that when initial length increases, total time cost and converging iterations decrease gradually, while RMSE and MAE fluctuates quite little with various initial antennae lengths.

Considering both accuracy and time cost, we choose the proper initial antennae length as 2.0 for D1 and D2.

*2) Comparison of computational efficiency*

The lowest RMSE/MAE and Friedman rank of all test models on D1-2 are given in Table II. The time cost is given in Table III. We draw the following conclusions that:

a) A²BAS-based optimizer's prediction accuracy is the highest of all models. For example, it can be observed that A²BAS's RMSE on D2 equals to 0.8615, which is 8.97% lower than Adam's 0.9464, 6.77% lower than PLFA's 0.9241, and 0.05% lower than HPL's 0.8656, respectively.

b) A²BAS-based optimizer's accuracy gain is statistically significant when compared with its peers. Friedman test reveals the fact that this optimizer has large advantage in accuracy. From Table II, it's evident that A²BAS-based optimizer owns lowest Friedman Rank value among all.

c) An A²BAS-based optimizer's time cost is higher than PLFA model, while it's much lower than the Adam-based LFA model.

Based on the above results and analyses, we conclude that the A²BAS-based optimizer obtains superiority in terms of prediction accuracy at the cost of higher time complexity.

TABLE II. COMPARISON RESULTS IN RMSE/MAE, INCLUDING WIN/LOSS COUNTS STATISTIC AND FRIEDMAN TEST

| Dataset | Metric | Adam | PLFA | HPL | A²BAS |
|---|---|---|---|---|---|
| D1 | RMSE | 0.7901 | 0.7856 | 0.7850 | **0.7843** |
|    | MAE  | 0.6087 | 0.6063 | 0.5998 | **0.5988** |
| D2 | RMSE | 0.9464 | 0.9241 | 0.8656 | **0.8615** |
|    | MAE  | 0.6661 | 0.6509 | 0.6394 | **0.6242** |
| Statistic | Win/Loss | 2/0 | 2/0 | 2/0 | 2/0 |
|    | Friedman Rank* | 4 | 3 | 2 | 1 |

* A lower Friedman rank value indicates higher rating prediction accuracy

TABLE III. COMPARISON RESULTS IN TIME COST

| Dataset | Adam | PLFA | HPL | A²BAS |
|---|---|---|---|---|
| D1 | 891 | 77 | 249 | 115 |
| D2 | 3393 | 48 | 222 | 331 |

## V. CONCLUSION

This paper focuses on refining the PSO-related LFA model with an innovative BAS algorithm. We design an A²BAS- model with two sub-algorithms. The first refinement sub-algorithm refines each row and column latent factor in a PLFA model to achieve a higher prediction accuracy. The refinement algorithm is the improved BAS algorithm which updates both beetles' antennae and step-size with Adam. The second sub-algorithm optimizes all the latent factors sequentially and updates the target HDI matrix. The experimental results from industrial datasets verify its high accuracy. In future, we plan to investigate a BAS swarm algorithm to refine the latent factors for HDI matrices.

REFERENCES

[1] X. Luo, Z. Liu, L. Jin, Y. Zhou and M. Zhou, "Symmetric Nonnegative Matrix Factorization-Based Community Detection Models and Their Convergence Analysis," in IEEE Transactions on Neural Networks and Learning Systems, vol. 33, no. 3, pp. 1203-1215, March 2022.
[2] D. Wu, X. Luo, M. Shang, Y. He, G. Wang and X. Wu, "A Data-Characteristic-Aware Latent Factor Model for Web Services QoS Prediction," in IEEE Transactions on Knowledge and Data Engineering.
[3] D. Wu, Q. He, X. Luo, M. Shang, Y. He and G. Wang, "A Posterior-Neighborhood-Regularized Latent Factor Model for Highly Accurate Web Service QoS Prediction," in IEEE Transactions on Services Computing, vol. 15, no. 2, pp. 793-805, 1 Mar.-Apr. 2022.
[4] X. Ma, D. Di, "Evolutionary nonnegative matrix factorization
[5] algorithms for community detection in dynamic networks," IEEE Trans. on Knowledge and Data Engineering, vol. 29, no. 5, pp. 1045-1058, May. 2017.
[6] S. Gao, H. Pang, P. Gallinari, J. Guo, and N. Kato, "A novel embedding method for information diffusion prediction in social network big data," IEEE Trans. on Industrial Informatics, vol. 13, no. 4, pp. 2097-2105, 2017.
[7] X. Luo, W. Qin, A. Dong, K. Sedraoui and M.-C Zhou, "Efficient and high-quality recommendations via momentum-incorporated parallel stochastic gradient descent-based learning," IEEE/CAA Journal of Automatica Sinica, vol. 8, no. 2, pp. 402-411,2021.
[8] X. Shi, Q. He, X. Luo, Y. Bai and M. Shang, "Large-Scale and Scalable Latent Factor Analysis via Distributed Alternative Stochastic Gradient Descent for Recommender Systems," in IEEE Transactions on Big Data, vol. 8, no. 2, pp. 420-431, 1 April 2022.
[9] D. Wu, M.-S Shang, X. Luo and Z.-D Wang, "An $L_1$-and-$L_2$-Norm-Oriented Latent Factor Model for Recommender Systems," IEEE Transactions on Neural Networks and Learning Systems.
[10] J. Chen, Y. Yuan, R. Tao, J. Chen, T. R. and X. Luo, "Hyper-parameter-evolutionary latent factor analysis for high-dimensional and sparse data from recommender systems," Neurocomputing, vol. 421, pp. 316-328, Jan. 2021.
[11] D. Chen, S. Li, D. Wu and X. Luo, "New disturbance rejection constraint for redundant robot manipulators: an optimization perspective," IEEE Transactions on Industrial Informatics, vol. 16, no. 4, pp. 2221-2232, 2020.
[12] L. Hu, X. Yuan, X. Liu, S. Xiong and X. Luo, "Efficiently Detecting Protein Complexes from Protein Interaction Networks via Alternating Direction Method of Multipliers," IEEE/ACM Transactions on Computational Biology and Bioinformatics, vol. 16, no. 6, pp. 1922-1935, Nov.-Dec. 2019.
[13] Z. Xie, L. Jin, X. Luo, S. Li and X. Xiao, "A data-driven cyclic-motion generation scheme for kinematic control of redundant manipulators," IEEE Transactions on Control Systems Technology, vol. 29, no. 1, pp. 53-63, 2020.
[14] X. Luo, Z. You, S. Li, Y. Xia and H. Leung, "Improving network topology-based protein interactome mapping via collaborative filtering," Knowledge-Based Systems, vol. 90, pp. 23-32, 2015.
[15] H. Lu, L. Jin, X. Luo, B. Liao, D. Guo and L. Xiao, "RNN for solving perturbed time-varying underdetermined linear system with double bound limits on residual errors and state variables," IEEE Trans. on Industrial Informatics, vol. 15, no. 11, pp. 5931-5942, Nov. 2019.
[16] Z. Liu, X. Luo, S. Li and M. Shang, "Accelerated Non-negative Latent Factor Analysis on High-Dimensional and Sparse Matrices via Generalized Momentum Method," 2018 IEEE International Conference on Systems, Man, and Cybernetics (SMC), 2018, pp. 3051-3056.
[17] D. Wu and X. Luo, "Robust latent factor analysis for precise representation of high-dimensional and sparse data.," IEEE/CAA Journal of Automatica Sinica, vol. 8, no. 4, pp. 796-805, 2020.
[18] X. Shi, Q. He, X. Luo, Y. Bai and M. Shang, "Large-Scale and Scalable Latent Factor Analysis via Distributed Alternative Stochastic Gradient Descent for Recommender Systems," in IEEE Transactions on Big Data, vol. 8, no. 2, pp. 420-431, 1 April 2022.
[19] X. Luo, Y. Yuan, S. Chen, N.-Y Zeng, and Z.-D Wang, "Position-transitional particle swarm optimization-incorporated latent factor analysis," IEEE Transactions on Knowledge and Data Engineering.

[20] J. Chen, X. Luo and M.-C Zhou, "Hierarchical particle swarm optimization-incorporated latent factor analysis for large-scale incomplete matrices," IEEE Transactions on Big Data.

[21] J.-F Chen, Y. Yuan, R. Tao, J. Chen, T. R. and X. Luo, "Hyper-parameter-evolutionary latent factor analysis for high-dimensional and sparse data from recommender systems," Neurocomputing, vol. 421, pp. 316-328, Jan. 2021.

[22] X. Luo, MC. Zhou, S. Li, Y. Xia, ZH. You, Q. Zhu, and H. Leung, "Incorporation of efficient second-order solvers into latent factor models for accurate prediction of missing QoS data," IEEE transactions on cybernetics, vol.48, no. 4, pp. 1216-1228, 172, 2017.

[23] X. Luo, H. Wu, H. Yuan and MC. Zhou, "Temporal pattern-aware QoS prediction via biased non-negative latent factorization of tensors," IEEE transactions on cybernetics, vol. 50, no.5, pp. 1798-1809, 2019.

[24] X. Luo, Z. Liu, S. Li, M. Shang and Z. Wang, "A fast non-negative latent factor model based on generalized momentum method," IEEE Transactions on Systems, Man, and Cybernetics Systems, vol. 51, no. 1, pp. 610-620, 99, 2018.

[25] Bottou, Léon. "Large-scale machine learning with stochastic gradient descent." Proceedings of COMPSTAT'2010. Physica-Verlag HD, pp. 177-186, 2010

[26] J. Duchi, E. Hazan, and Y. Singer, "Adaptive subgradient methods for online learning and stochastic optimization," Journal of Machine Learning Research, vol. 12, no. 7, pp. 2121-2159, 2011.

[27] Zeiler, M. D., "ADADELTA: An Adaptive Learning Rate Method", arXiv e-prints, 2012.

[28] Kingma, D. P. and Ba, J., "Adam: A Method for Stochastic Optimization", arXiv e-prints, 2014.

[29] X. Jiang, and S. Li, "BAS: Beetle Antennae Search Algorithm for Optimization Problems." International Journal of Robotics and Control 1.1, 2017.

[30] T. Wang, Y. Long, "Beetle swarm optimization algorithm: Theory and application," arXiv e-prints, 2018.

[31] H. Zhang, Z. Li, X. Jiang, et al. "Beetle colony optimization algorithm and its application," IEEE Access, vol. 8, pp. 128416-128425, 2020.

[32] A. H. Khan, X. Cao, S. Li, V. N. Katsikis and L. Liao, "BAS-ADAM: an ADAM based approach to improve the performance of beetle antennae search optimizer," in IEEE/CAA Journal of Automatica Sinica, vol. 7, no. 2, pp. 461-471, March 2020.

[33] J. Konstan, B. Miller, D. Maltz, J. Herlocker, L. Gordon, and J. Riedl, "Grouplens: applying collaborative filtering to usenet news," Communications of the ACM, vol. 40, no. 3, pp. 77-87, 1997.

[34] J. Mohsen and E. Martin, "A matrix factorization technique with trust propagation for recommendation in social networks," in Proc. of the 4th ACM Conf. on Recommender Systems, Barcelona, Spain, 2010, pp. 135-142

[35] J. Chen, R. Wang, D. Wu and X. Luo, "A Differential Evolution-Enhanced Position-Transitional Approach to Latent Factor Analysis," IEEE Transactions on Emerging Topics in Computational Intelligence, 2022, doi: 10.1109/TETCI.2022.3186673.